# ADAPTIVE MULTI-FIDELITY REINFORCEMENT LEARNING FOR VARIANCE REDUCTION IN ENGINEERING DESIGN OPTIMIZATION

**Akash Agrawal**
Mechanical Engineering
Carnegie Mellon University
Pittsburgh, PA, USA
aragrawa@alumni.cmu.edu

**Christopher McComb**
Mechanical Engineering
Carnegie Mellon University
Pittsburgh, PA, USA
ccm@cmu.edu

## ABSTRACT

*Multi-fidelity Reinforcement Learning (RL) frameworks efficiently utilize computational resources by integrating analysis models of varying accuracy and costs. The prevailing methodologies, characterized by transfer learning, human-inspired strategies, control variate techniques, and adaptive sampling, predominantly depend on a structured hierarchy of models. However, this reliance on a model hierarchy can exacerbate variance in policy learning when the underlying models exhibit heterogeneous error distributions across the design space. To address this challenge, this work proposes a novel adaptive multi-fidelity RL framework, in which multiple heterogeneous, non-hierarchical low-fidelity models are dynamically leveraged alongside a high-fidelity model to efficiently learn a high-fidelity policy. Specifically, low-fidelity policies and their experience data are adaptively used for efficient targeted learning, guided by their alignment with the high-fidelity policy. The effectiveness of the approach is demonstrated in an octocopter design optimization problem, utilizing two low-fidelity models alongside a high-fidelity simulator. The results demonstrate that the proposed approach substantially reduces variance in policy learning, leading to improved convergence and consistent high-quality solutions relative to traditional hierarchical multi-fidelity RL methods. Moreover, the framework eliminates the need for manually tuning model usage schedules, which can otherwise introduce significant computational overhead. This positions the framework as an effective variance-reduction strategy for multi-fidelity RL, while also mitigating the computational and operational burden of manual fidelity scheduling.*

## 1. INTRODUCTION

Multi-fidelity Reinforcement Learning (RL) has become a powerful strategy for accelerating engineering design by strategically combining simulation models of varying fidelity, ranging from simplified surrogates to detailed, computationally demanding analyses [1–5]. High-fidelity simulations, such as computational fluid dynamics or finite element analysis, offer precision but at considerable computational cost. Conversely, low-fidelity models, including simplified physics or data-driven surrogates, provide rapid yet less accurate insights into system behavior [6–10]. By effectively combining these multiple fidelities, RL algorithms significantly improve computational efficiency, enabling practical application of RL in resource-constrained engineering design problems.

However, current multi-fidelity RL frameworks rely predominantly on hierarchical model sequences with fixed fidelity schedules [1–5]. While these hierarchical structures streamline model management, they inadvertently introduce significant variance in policy learning. Specifically, rigid schedules of model usage overlook the heterogeneity of model errors across different regions of the design space. Such heterogeneity emerges due to variations in modeling assumptions [6–9], model complexity mismatch with design problem [11–13], diverse data availability [14–17], and challenges posed by hybrid modeling techniques [18,19]. Consequently, RL algorithms employing rigid hierarchies often experience substantial fluctuations in policy updates and poor convergence, limiting their practical utility in complex engineering problems.

Reducing variance in policy learning is essential to ensure consistent high-quality solutions to engineering design problems. Specifically, variance refers to uncertainty or inconsistency in policy updates arising from noisy or misaligned data across fidelity levels, adversely impacting the reliability of RL-driven design. Motivated by this perspective, the present work proposes a novel adaptive multi-fidelity RL framework designed to reduce variance through adaptive policy learning. Unlike conventional hierarchical methods, our proposed framework adaptively leverages multiple heterogeneous, non-hierarchical low-fidelity models alongside a high-fidelity model to efficiently learn a high-fidelity policy. Specifically, low-fidelity policies and their experience data are adaptively used for efficient targeted learning, guided by their alignment with the high-fidelity policy. This alignment, measured through cosine similarity of policy action means, ensures correlated sampling and systematically reduces the variance inherent to policy learning. By adaptively managing the selection of fidelity models according to local alignment rather than globally imposed hierarchies, the proposed approach decreases variance in learning and consistently achieves high-quality solutions.

The remainder of this paper is structured as follows. Section 2 briefly contextualizes multi-fidelity RL, highlighting key challenges of variance in context of heterogeneous analysis models. Section 3 introduces the proposed adaptive RL approach as a variance-reduction framework, describing its alignment-based strategy in detail. Section 4 presents an engineering case study on octocopter design optimization to demonstrate the effectiveness of the framework in practical use. Section 5 presents results demonstrating significant improvements in variance reduction, which lead to enhanced convergence and consistent high-quality solutions compared to traditional hierarchical multi-fidelity methods. Section 6 summarizes this work's primary contributions and proposes future research directions to further enhance the adaptivity and efficiency of multi-fidelity RL.

## 2. BACKGROUND

Reinforcement learning [20–23] has emerged as a powerful computational paradigm for adaptively exploring complex, non-differentiable, multi-dimensional, dynamic design spaces characterized by multiple objectives and constraints [24–38]. However, in engineering design, practical RL implementation faces significant computational bottlenecks due to the high cost of detailed, high-fidelity simulations (such as computational fluid dynamics or finite element analyses) that compose the reward function. Multi-fidelity RL frameworks directly address this computational challenge by leveraging combinations of high-fidelity simulations with computationally efficient, lower-fidelity models. Low-fidelity models such as surrogate models, reduced-order approximations, simplified physics simulations, and partially converged results [6–10], offer rapid, approximate analyses at significantly reduced computational costs. State-of-the-art multi-fidelity RL approaches typically employ strategies including human-inspired methods [1] transfer learning [2,3], control variate techniques [4], and adaptive sampling [5], to enhance efficiency. However, these prevailing approaches generally depend on hierarchical structures, which assume a monotonic progression of fidelity levels. Within these hierarchical frameworks, lower-fidelity models provide broad preliminary searches, while progressively higher-fidelity models are employed at later stages for refined evaluations.

Despite their computational appeal, hierarchical frameworks often overlook significant sources of variance that emerge from heterogeneous error distributions of lower-fidelity models across different regions of the design space [10,39,40]. Variance in RL policy learning arises primarily because the effectiveness of lower-fidelity models as proxies for high-fidelity simulations can vary significantly depending on local design contexts [11–13]. For instance, the geometry of a design significantly influences the dominance of laminar or turbulent flow. This variation in flow type across different regions affects model accuracy because different models are typically calibrated for specific flow conditions. A model that performs well under laminar flow might not accurately capture the complexities of turbulent flow. Therefore, using a model that does not account for the dominant flow type in a particular region of the design space can lead to errors. Moreover, different low-fidelity models might adopt varying assumptions about the flow type and employ distinct methods to model that flow type, impacting model accuracy.

Surrogate models trained from limited or unevenly distributed datasets similarly produce heterogeneity in error distributions, often failing to generalize well across the design space [14–17]. Furthermore, datasets with partially converged results of high-fidelity simulations may also lead to heterogeneity in model errors. Hybrid models combine complementary approaches (e.g., empirical data with physics-based simulations [19]) to enhance accuracy. However, imperfect complementarity can lead to spatially varying accuracy and increased complexity due to unpredictable interactions among model components with distinct error characteristics [18]. Such heterogeneity introduces inconsistent signals during policy updates, substantially increasing variance, thus impairing the reliability of the RL agent to yield high-quality design solutions.

Recent research has increasingly recognized the detrimental impacts of variance in policy learning and explored explicit variance reduction techniques. Khairy and Balaprakash [4] introduced a control variate method in multi-fidelity RL that taps into the correlation between low and high-fidelity returns, leading to a reduced variance in $Q$ estimates and a better resultant policy. In another line of work by Qiu et al. [41], they proposed a multi-fidelity simulator framework for multi-agent RL, utilizing depth-first search strategies on low-fidelity simulators to derive expert local policies. These expert policies effectively guide high-fidelity exploration, reducing variance and significantly lowering computational costs in multi-agent navigation tasks.

This work specifically builds on an earlier hierarchical framework [1] that progressively utilizes models from low- to high-fidelity within episodic design tasks to achieve a high solution quality at a reduced computational cost. In the current work, we extend the framework to adaptively handle heterogeneous, non-hierarchical low-fidelity models alongside a high-fidelity one to learn a high-fidelity policy, with the goal of reducing variance arising from heterogeneous model errors.

## 3. METHODOLOGY

This work proposes an adaptive multi-fidelity RL framework designed to reduce variance in policy learning by adaptively leveraging heterogeneous low-fidelity (*LF*) models alongside a high-fidelity (*HF*) model. This section details the methodology used to construct and evaluate the proposed framework. Section 3.1 introduces the adaptive multi-fidelity framework emphasizing variance reduction via an alignment metric between *LF* and *HF* policies. Section 3.2 describes the training and evaluation process, explicitly assessing variance in solution quality and computational efficiency relative to hierarchical and single-fidelity RL baselines.

## 3.1 Adaptive Multi-fidelity RL Framework

The adaptive multi-fidelity RL framework developed in this work extends a prior multi-fidelity RL approach [1] to reduce variance arising from heterogeneous model errors during policy learning. Like previous methods, our framework solves an engineering design optimization problem by starting from initial seed designs and iteratively optimizing continuous and discrete design variables to minimize an objective $f'$, subject to constraints $G'$ and $H'$. Consistent with prior approaches, we assume explicit knowledge of the highest fidelity level.

In contrast to traditional hierarchical frameworks, which rely on fixed fidelity sequences, our approach assigns a dedicated RL agent to each fidelity model (one high-fidelity agent and multiple low-fidelity agents), explicitly recognizing heterogeneity across model error distributions. Each agent learns a policy ($\pi$) for generating actions ($a_t$) in response to states ($s_t$), with the goal of maximizing cumulative rewards derived from its respective analysis model. Rewards ($r$) are formulated based on objective improvements and constraint satisfaction, as in the previous multi-fidelity RL method [1]. The adaptive interaction of the agents with the design space is illustrated conceptually in Figure 1 and detailed algorithmically in Algorithm 1.

The adaptive multi-fidelity RL framework comprises one high-fidelity ($HF$) agent and multiple low-fidelity ($LF_i$) agents. These agents interact with the design space environment, collecting experience data used to update their respective policies. This experience data contains states, actions, and rewards observed during interactions with the design space. During training, variance in $HF$ policy learning can arise from inconsistent policy updates if all $LF$ experience data (with underlying heterogeneous error distributions) is directly integrated into the training of the $HF$ policy. Such variance can ultimately affect convergence and the quality of design solutions. To mitigate this variance, the proposed framework adaptively prioritizes $LF$ data usage based on the alignment of $LF$ policies with the $HF$ policy. Specifically, alignment is quantitatively assessed using the cosine similarity between the mean action distributions of the low-fidelity policies ($\overline{\pi}_{LF_i}$) and the high-fidelity policy ($\overline{\pi}_{HF}$). It is important to emphasize that this alignment metric assesses policy similarity at each step based on the current, evolving state of the policies.

The case of two $LF$ agents ($LF_1$ and $LF_2$) is illustrated in Figure 1 and discussed in further detail here. Specifically, at each step, an alignment threshold (illustrated as 45 degrees in Figure 1) determines whether each $LF$ policy aligns sufficiently with the $HF$ policy. Accordingly, the adaptive alignment measure yields four possible scenarios at each training step: both $LF_1$ and $LF_2$ align with $HF$; $LF_1$ aligns, but $LF_2$ does not; $LF_2$ aligns, but $LF_1$ does not; neither $LF_1$ nor $LF_2$ align with $HF$. In regions where $LF$ policies closely align with the $HF$ policy, the $HF$ agent leverages the aligned $LF$ agents' experience data for policy updates. Essentially, rewards from $LF$ models are used in aligned portions of a trajectory, and $HF$ value estimates are used for beyond. This adaptive integration enables efficient targeted learning, effectively reducing computational costs without inducing variance in learning. Conversely, in critical regions where $LF$ agents diverge substantially from the $HF$ policy, indicating a source of potentially high variance, the $HF$ agent itself is utilized directly, ensuring precision in policy learning.

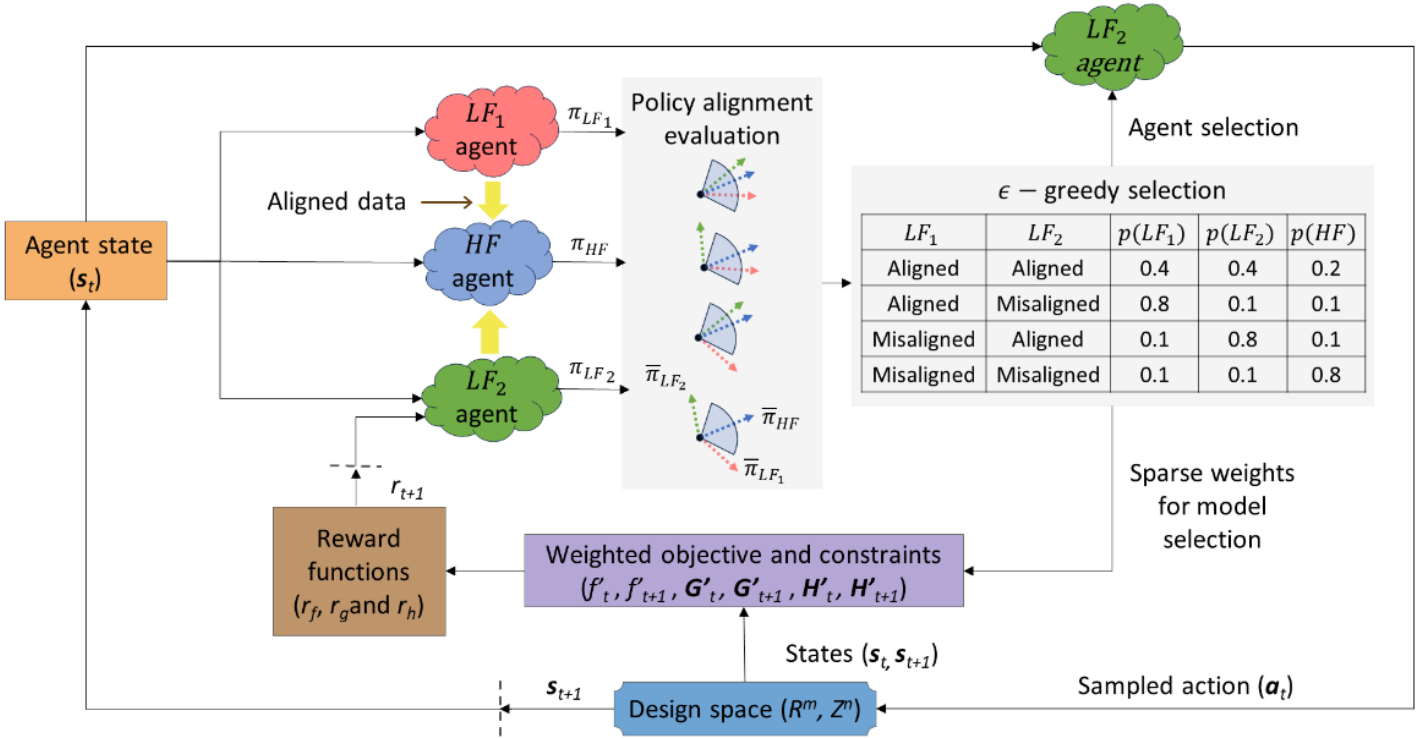

**FIGURE 1: ADAPTIVE MULTI-FIDELITY RL FRAMEWORK**

The specific technique for model fidelity selection at each iteration is an $\epsilon$-greedy strategy (Algorithm 2), balancing exploration of fidelity options and exploitation of existing alignment knowledge. This strategy prevents premature fixation on a particular fidelity model, ensuring each region in the design space receives adequate exploration across fidelities. Once a fidelity is selected (e.g., $LF_2$ in Figure 1), the chosen agent samples an action from its policy, evaluates the associated objectives and constraints, and computes the resulting reward, thus completing a timestep.

Furthermore, the alignment threshold dynamically evolves during training using a cosine schedule [42], progressively adjusting from an initial broad alignment criterion (90 degrees) toward stricter alignment (0 degrees). This schedule allows early reliance on computationally inexpensive low-fidelity models while transitioning toward increased usage of the high-fidelity model in later training stages. Such an approach systematically balances computational efficiency and policy accuracy over time, limiting variance in the policy learning process.

The alignment assessment based on the evolving state of the policies at each iteration ensures that the integrated low-fidelity experiences consistently provide correlated reliable signals. This granular adaptive alignment serves as an effective guide to the ongoing high-fidelity policy learning process, systematically limiting variance while enhancing computational efficiency.

### 3.2 Training and Evaluating RL Agents

Agents in the proposed adaptive multi-fidelity framework are trained using the proximal policy optimization (PPO) algorithm [43], employing multiple randomly sampled seed designs. Upon completion of training, the learned $HF$ policy is evaluated using the same seed designs, measuring solution quality exclusively with the high-fidelity model. To better contextualize the effectiveness of our adaptive variance-reduction strategy, we perform a comparative study against the hierarchical multi-fidelity RL framework [1] from prior work, along with individual single-fidelity RL baselines trained separately using each fidelity model ($HF$, $LF_1$ and $LF_2$). For the hierarchical framework, two specific configurations are tested to investigate how the sequence of $LF$ models affect variance in the learning process, each employing a predefined percentage of

---

*Algorithm 1: Adaptive Multi − fidelity RL*

$\pi_{HF}, \pi_{LF_1}, \pi_{LF_2} \leftarrow \pi_0$ (policy functions)
$v_{HF}, v_{LF_1}, v_{LF_2} \leftarrow v_0$ (value functions)
$B_{HF}, B_{LF_1}, B_{LF_2} \leftarrow [\,]$ (data buffers)
$\epsilon \coloneqq constant$ (Model choice $\epsilon$-greedy parameter)

**for** $e \subset \{0, 1, \ldots, EPISODE\ COUNT - 1\}$ **do**
  $s_0 \leftarrow$ sample seed design
  **for** $t \subset \{0, 1, \ldots, EPISODE\ LENGTH - 1\}$ **do**
    $S_{cos,1} \leftarrow$ cosine similarity between $\overline{\pi}_{LF_1}(s_t)$ and $\overline{\pi}_{HF}(s_t)$
    $S_{cos,2} \leftarrow$ cosine similarity between $\overline{\pi}_{LF_2}(s_t)$ and $\overline{\pi}_{HF}(s_t)$
    $model_t, aligned_t \leftarrow$ **model choice** $(e, S_{cos,1}, S_{cos,2}, \epsilon)$
    $a_t \leftarrow$ sample action from $\pi_{model_t}(s_t)$
    $s_{t+1}, r_{t+1} \leftarrow step_{model_t}(s_t, a_t)$
    **if** $t > 0$ AND $model_t = model_{t-1}$
      $B_{model_t} \ll (s_t, a_t, r_{t+1}, s_{t+1}, aligned_t)$
      (append to current sequence)
    **else**
      $B_{model_t} += [(s_t, a_t, r_{t+1}, s_{t+1}, aligned_t)]$
      (start new sequence)
    $s_t \leftarrow s_{t+1}$
  **end**

  Augment $HF$ data with $LF$ data where there is alignment
  **for** $model$ in $LF_1, LF_2$ **do**
    **for** $sequence$ in $B_{model}$ **do**
      $subsequences \leftarrow$ find contiguous subsets in $sequence$ where $aligned = True$
      $B_{HF} += [subsequence$ **for** $subsequence$ in $subsequences]$
    **end**
  **end**

  **for** $model$ in $LF_1, LF_2, HF$ **do**
    **if** $BATCH$ data collected in $B_{model}$
      train $\pi_{model}$ and $v_{model}$ using $BATCH$ and policy opt. algo.
      $B_{model} \leftarrow [\,]$
  **end**
**end**

---

*Algorithm 2: Model choice*
**function** model choice $(e, S_{cos,1}, S_{cos,2}, \epsilon)$

$$\theta \leftarrow \begin{cases} \cos\left[\frac{\pi}{4}\left(1 + \cos\left(\pi \cdot \frac{e}{0.9 \times EP_{MAX}}\right)\right)\right], & e < 0.9 \times EP_{MAX} \\ 0, & e \geq 0.9 \times EP_{MAX} \end{cases}$$

$$(p_{LF_1}, p_{LF_2}, p_{HF}) \leftarrow \begin{cases} \left(\frac{1-\epsilon}{2}, \frac{1-\epsilon}{2}, \epsilon\right), & S_{cos,1} > \theta \text{ AND } S_{cos,2} > \theta \\ \left(1-\epsilon, \frac{\epsilon}{2}, \frac{\epsilon}{2}\right), & S_{cos,1} > \theta \text{ AND } S_{cos,2} < \theta \\ \left(\frac{\epsilon}{2}, 1-\epsilon, \frac{\epsilon}{2}\right), & S_{cos,1} < \theta \text{ AND } S_{cos,2} > \theta \\ \left(\frac{\epsilon}{2}, \frac{\epsilon}{2}, 1-\epsilon\right), & S_{cos,1} < \theta \text{ AND } S_{cos,2} < \theta \end{cases}$$

$$model \leftarrow Categorical(\{LF_1, LF_2, HF\}, \{p_{LF_1}, p_{LF_2}, p_{HF}\})$$

$$aligned = \left[model == (\underset{m \subset \{LF_1, LF_2, HF\}}{\arg\max} p_m)\right]$$

**return** $model, aligned$

steps for $LF_1$, $LF_2$, and $HF$ models. These configurations are (35% $LF_1$, 35% $LF_2$, 30% $HF$) and (35% $LF_2$, 35% $LF_1$, 30% $HF$). We aimed to allocate equal training time to each fidelity level, but an exact equal split was infeasible due to the total number of steps in the episode being 20. Therefore, we adjusted the allocation to 35% of the steps for $LF_{1\ or\ 2}$ (7 steps), 35% for $LF_{2\ or\ 1}$ (7 steps), and 30% for $HF$ (6 steps).

To assess the adaptive capability of our approach, model usage is tracked throughout the training process. Additionally, to evaluate variance reduction achieved by the adaptive alignment-based framework, we analyze the spread of solution quality across different methods. Moreover, we assess computational efficiency by analyzing the total computational cost of evaluating objectives and constraints.

## 4. OCTOCOPTER DESIGN CASE STUDY

In order to evaluate the proposed adaptive multi-fidelity RL framework, we demonstrate its application on an octocopter design problem. This problem involves a corpus of components and a flight dynamics simulator [44,45]. The components include batteries, motors, and propellers. The design space of the problem comprises a continuous variable for arm length and three ordinal variables for the choice of batteries, motors, and propellers from an ordered set of components. Figure 2 illustrates the design artifact of an octocopter generated by assigning random values to the design variables. The reader is referred to prior work [45] for details on the corpus of components used in this problem. By considering all possible discrete values and merely 10 values for the continuous variable, the size of the combinatorial space is of the order of $10^6$.

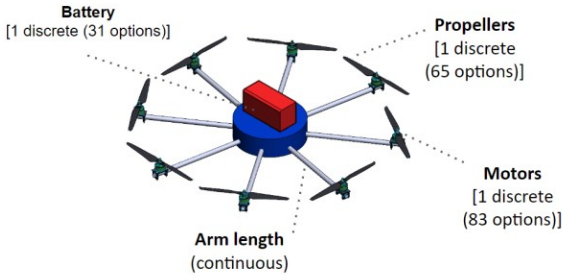

**FIGURE 2: DESIGN ARTIFACT OF OCTOCOPTER**

**(LABELS INDICATE THE DESIGN VARIABLES ASSOCIATED WITH DIFFERENT COMPONENTS)**

The design objective is based on a maneuvering task along a trajectory defined by a set of waypoints. Specifically, we define a maximization objective as follows:

$$Q = \frac{d}{D} \times \bar{v} \times (1 - \bar{e}) \quad (1)$$

where $d$ is the distance covered along the trajectory, $D$ is total distance to be covered to complete the trajectory, $\bar{v}$ is the normalized average speed of the maneuver, and $\bar{e}$ is the normalized average error (deviation) from the specified trajectory. To emphasize, this objective aims at developing long-range, fast and stable quadcopters. Figure 3 showcases the simulated trajectories for two exemplar octocopter configurations. Specifically, the components of each octocopter are listed. Further, the path followed by them during the simulation is visualized in three orthogonal planes along with the reference path. The reader is referred to prior work [44] on the flight dynamics simulator for further details.

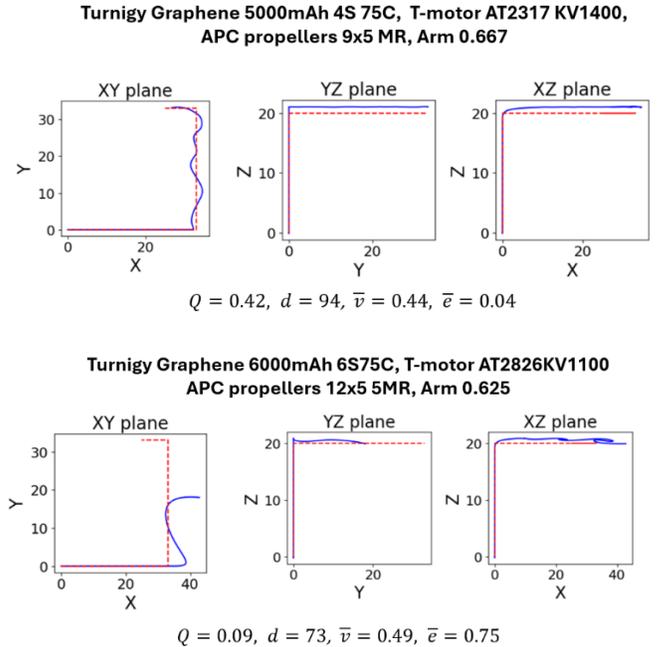

**FIGURE 3: SIMULATION OF EXEMPLAR OCTOCOPTERS**

The flight dynamics simulator serves as a high-fidelity ($HF$) model for this case study. The median cost of this model on an Intel Xeon CPU used in this work is 1.78 s. Additionally, the cost at the 25th percentile is 0 s, while at the 75th percentile it is 22.04 s, indicating significant variability in the range and speed of the octocopters. Notably, the zero second values represent octocopters that are not flyable due to interferences or incompatibility of the components. The low-fidelity models ($LF_1, LF_2$) are prepared by training two neural networks on distinct datasets obtained from a partially converged optimization run for designing the octocopter. This optimization data was scaled using a min–max normalization technique. Figure 4 shows the reduced dimensional Principal Component Analysis (PCA) space of this partially converged scaled data with the highlighted subsets reflecting the datasets used to train the low-fidelity models. The tailoring of each model to different regions of the search space leads to heterogenous models, contrasting with a rigid hierarchy and providing an effective testbed for the proposed framework.

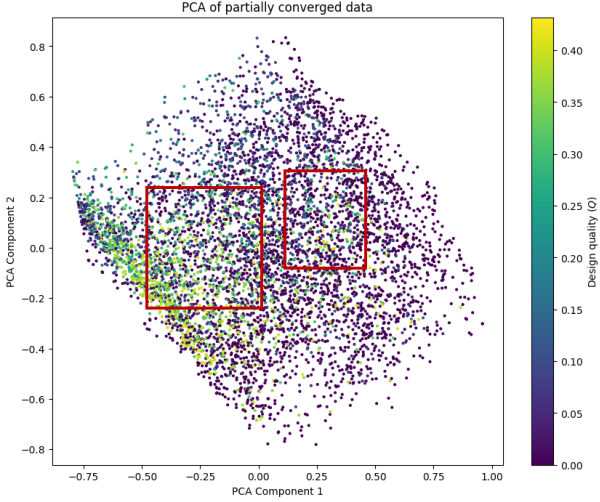

**FIGURE 4: DATASETS USED TO TRAIN LOW-FIDELITY SURROGATES**

The architecture of both the $LF_1$ and $LF_2$ neural networks is defined as follows:

$$I_4 - D_{64,R} - D_{32,R} - O_{1,L}$$

where $I_{N_i}$ represents the input layer of size $N_i$, $D_{j,k}$ represents a dense (hidden) layer of size $j$ with an activation denoted by $k$, $R$ represents the ReLU activation, $L$ represents linear activation, and $O_{N_o,k}$ represents the output layer of size $N_o$ with activation k. The size of the datasets used to train the $LF_1$ and $LF_2$ models are 680 and 1358 respectively. The prediction accuracies of the $LF_1$ and $LF_2$ models on the validation subsets of their respective datasets are 0.51 and 0.56, respectively. Furthermore, their prediction accuracies on the entire dataset shown in Figure 4 are 0.23 and 0.38, respectively. This implies that the low-fidelity models are indeed tailored to different regions of the search space. Lastly, the mean cost of evaluation for both these models on an Intel Xeon CPU used in this work is 208 µs.

## 5. RESULTS AND DISCUSSION

The RL policies were trained and evaluated with the proposed adaptive multi-fidelity RL framework, a hierarchical multi-fidelity RL framework, and with each of the models individually. Further, 1200 seed points and episodes were utilized for training and evaluating all the policies. The results of the evaluation for the six cases (Adaptive MFRL, Hierarchical MFRL 1, Hierarchical MFRL 2, High-fidelity RL, Low-fidelity 1 RL, and Low-fidelity 2 RL) as per the high-fidelity model are shown in Figure 5.

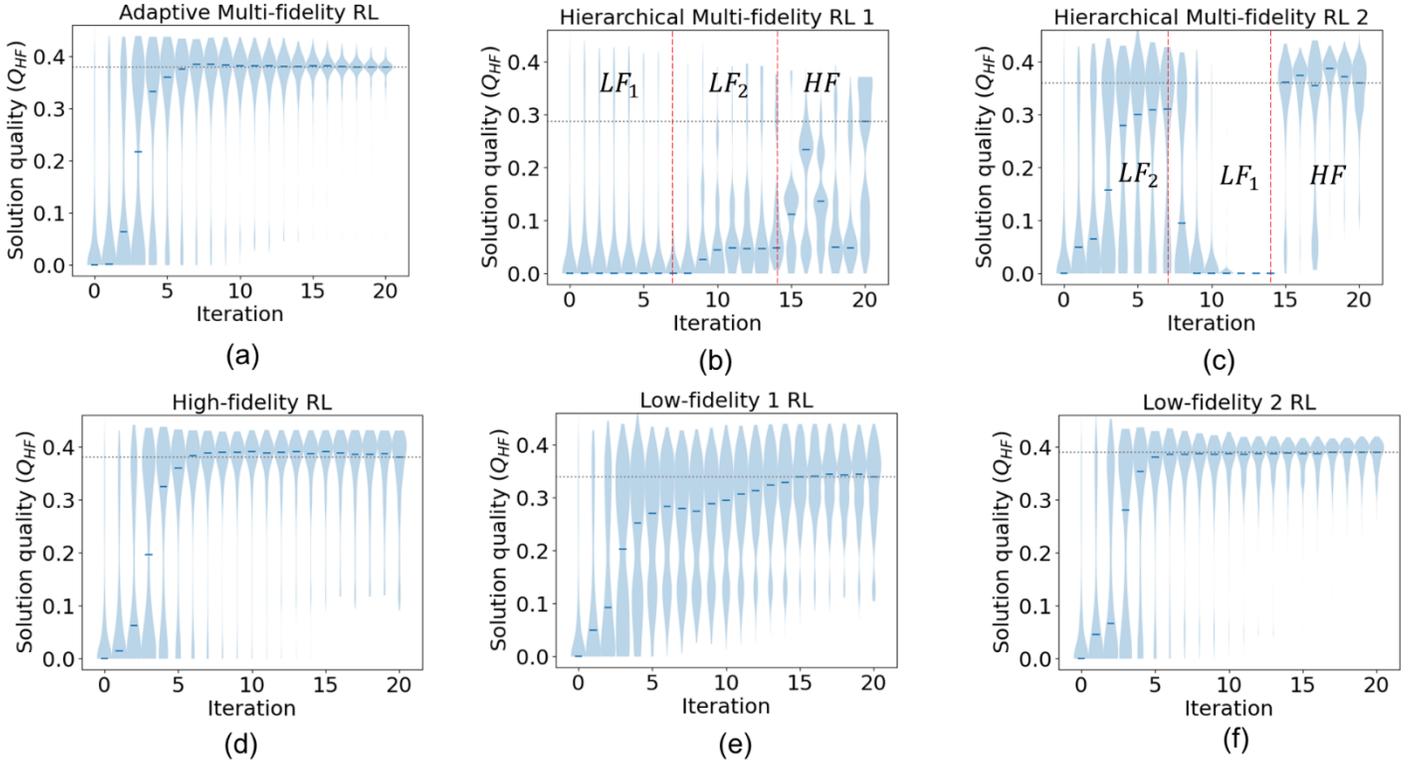

**FIGURE 5: TRAINED POLICIES ARE EVALUATED BY PASSING SEEDS**

The quality of the solutions for all the cases is better than the seed designs, albeit with drastically different trends across the agents. For the Adaptive MFRL case (Fig. 5(a)), there is a high dispersion in initial iterations followed by convergence to a high quality with low variance. This indicates that the model can consistently find high-quality solutions. For the Hierarchical MFRL 1 case (Fig. 5(b)), the quality does not improve when the $LF_1$ model is operational. Further, the quality rises slowly when the $LF_2$ model is operational. Lastly, a steep rise to a moderate quality is observed when the agent switches to the high-fidelity model. However, this rise is not consistent across the iterations of the episode. For the Hierarchical MFRL 2 case (Fig. 5(c)), the quality rises significantly with a high dispersion in initial iterations when the $LF_2$ model is operational. Further, after switching to the $LF_1$ model, the quality drops to a poor value. Lastly, the quality rises to a high value with the use of the $HF$ model similar to the adaptive case (Fig. 5(a)), albeit with a higher variance. These trends of the hierarchical cases indicate that the performance of the underlying framework is sensitive to the ordering as well as the proportions of model usage. This contrasts with the adaptive framework which is not restricted by a predefined schedule of model usage. For the High-fidelity RL case (Fig. 5(d)), we observe a trend similar to the Adaptive MFRL case (Fig. 5(a)) with a high solution quality, albeit with a higher variance. This is potentially because the agent has not fully learned the underlying complexities of the high-fidelity model within 1200 episodes. Lastly, for the cases that just utilize one of the low-fidelity models (Figs. 5(e) and 5(f)), the agents converge to high-quality solutions similar to the adaptive case (Fig. 5(a)). However, they both have higher variance than the adaptive case. In summary, the adaptive model stands out due to yielding both a high solution quality and minimal variance.

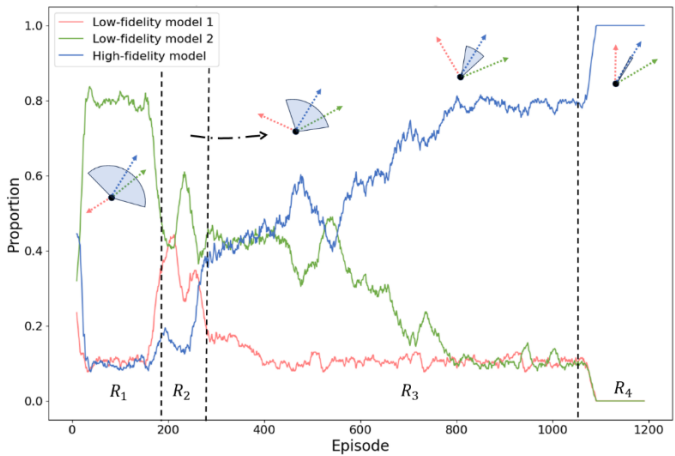

**FIGURE 6: MODELS ARE ADAPTIVELY CHOSEN ACROSS TRAINING TIME**

The proportion of usage of different models is also evaluated across time. Figure 6 shows the evolution of model usage across the training of the agent with four distinct regimes ($R_1, R_2, R_3, R_4$) in the trends. Initially in the regime $R_1$, the $LF_2$ model is heavily utilized, with its proportion quickly rising above 0.8. Meanwhile, the high-fidelity model and low-fidelity model 1 are used sparingly. This is potentially because the $LF_2$ policy is more aligned to the $HF$ policy than the $LF_1$ policy, even with a loose initial alignment threshold. As training progresses into regime $R_2$, the usage of both the low-fidelity models begins to fluctuate, essentially complementing each other for a few episodes. This is indicative of the varied alignment of the low-fidelity policies with the high-fidelity policy. In the regime $R_3$, there is a decrease in the usage of both low-fidelity models, with the decrease in the $LF_1$ model much more rapid the $LF_2$ model. This decrease coincides with an increase in the proportion of the $HF$ model. This transition suggests that as the alignment threshold tightens the algorithm relies on more accurate feedback from the $HF$ model. Moreover, the different rates of decrease of the $LF$ models correspond to the varied alignment of the $LF$ policies with respect to the $HF$ policy. The last regime $R_4$ serves to increasingly refine the learning of the $HF$ policy with negligible influence of the $LF$ models.

To better contextualize the effectiveness of the adaptive multi-fidelity RL framework, we examine the variance in solution quality and the associated computational efficiency compared to hierarchical and single-fidelity baselines.

Figure 7(a) illustrates the distribution of solution quality across agents, revealing notable differences in variance. While all agents generally achieve high-performance solutions on average, substantial variance is observed in the Low-fidelity 1 RL and both Hierarchical MFRL configurations, indicating inconsistent policy learning and reduced reliability. The High-fidelity RL agent also yields high variance in quality, potentially because it has not fully learned the underlying complexities of the high-fidelity model within the limited number of episodes. The Low-fidelity 2 RL agent demonstrates lower variance than the aforementioned cases, reflecting relatively more stable performance; however, it still shows greater variance compared to the Adaptive MFRL agent. Notably, the Adaptive MFRL agent uniquely demonstrates significantly reduced variance relative to all other agents, consistently achieving high-quality solutions due to adaptive alignment-based policy updates. This distinctly lower variance highlights the robustness and stability uniquely afforded by the adaptive framework.

In terms of computational efficiency, illustrated in Figure 7(b), the purely low-fidelity agents incur minimal computational costs. The hierarchical approaches incur moderate computational expenses, but their solution quality depends significantly on the chosen ordering of fidelity models, potentially requiring additional computational effort and time to systematically explore multiple configurations. In contrast, although the adaptive agent involves higher computational expense compared to low-fidelity and individual hierarchical configurations, it reduces variance and achieves reliable convergence without the need for explicitly exploring model orderings and proportions. Lastly, the purely $HF$ agent incurs the highest computational costs and would require significantly extended training episodes to achieve comparable stable convergence, further increasing its computational expense.

Thus, the adaptive multi-fidelity framework prioritizes variance reduction, ensuring reliable policy learning, while still effectively managing computational expenses without explicit model scheduling.

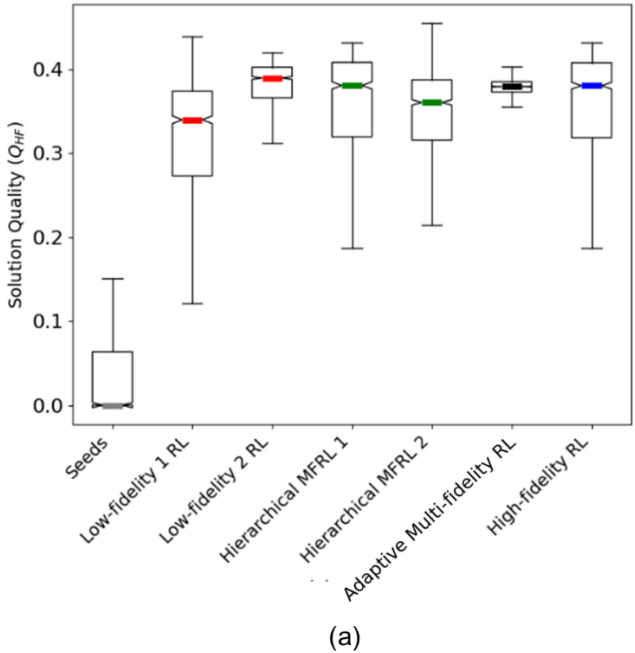

(a)

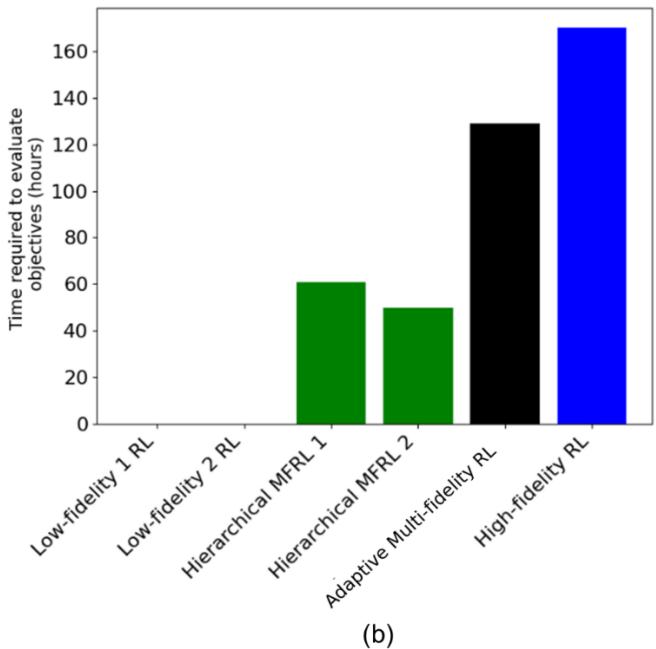

(b)

**FIGURE 7: SOLUTION QUALITY AND EFFICIENCY COMPARISONS WITH BASELINES**

## 6. CONCLUSION

This research presents a novel adaptive multi-fidelity reinforcement learning framework designed to reduce variance in policy learning by adaptively utilizing heterogeneous low-fidelity models alongside a high-fidelity model. In contrast to traditional hierarchical multi-fidelity methods, our adaptive framework dynamically selects models based on the alignment between low-fidelity policies and the high-fidelity policy, systematically reducing variance in policy updates. We demonstrate the effectiveness of this adaptive variance-reduction strategy through a case study involving an octocopter design problem. Results indicate that our framework uniquely achieves consistently high-quality solutions with significantly reduced variance, highlighting superior convergence relative to hierarchical baselines. Moreover, adaptive fidelity selection effectively balances variance reduction and computational efficiency without explicit model scheduling.

Future research should focus on further enhancements to the framework's adaptive mechanisms, exploring alternative alignment metrics and threshold scheduling techniques to achieve greater variance reduction and further improved computational efficiency. Investigating cost-weighted model selection probabilities when multiple low-fidelity policies align with the high-fidelity policy could additionally enhance the efficiency-variance tradeoff. While our current framework explicitly handles heterogeneous models without predefined hierarchies, incorporating partial hierarchical structures within some fidelity levels may further enhance performance by combining the advantages of hierarchical and adaptive strategies. Future work could also extend the framework's applicability across diverse representations and multi-disciplinary optimization scenarios [46], leveraging interactions between disciplines for even more robust variance-reducing strategies. Additionally, design space visualization techniques [47,48] could enhance explainability [49] by providing valuable insights into fidelity selection patterns, aiding in further refinement of adaptive learning strategies. Future studies could also leverage systematic testbeds such as those proposed by Tao et al. [50] to better evaluate the adaptive variance-reduction strategy under controlled and well-characterized multi-fidelity settings.

Comparative studies with RL-based methodologies explicitly modeling errors across fidelity levels, as well as investigations into other established multi-fidelity management strategies [6,7], will provide valuable insights into the adaptive framework's strengths and limitations. Finally, exploring budget-aware extensions to optimize resource allocation [51] will significantly increase the practical utility of the adaptive approach, enabling it to robustly tackle complex engineering design challenges across various resource-constrained domains.

## ACKNOWLEDGEMENTS

The authors are grateful to Srikanth Devanathan of Dassault Systèmes for his feedback on early versions of this work.